\pdfoutput=1
\documentclass[twoside,twocolumn]{article}
\usepackage{blindtext} 

\usepackage[sc]{mathpazo} 
\usepackage[T1]{fontenc} 
\usepackage{microtype} 

\usepackage{cite}
\usepackage{amsmath,amssymb,amsfonts}
\usepackage{algorithmic}
\usepackage{graphicx}
\usepackage{extarrows}
\usepackage{multirow}
\usepackage{threeparttable}
\usepackage{array}
\usepackage{amssymb}
\usepackage{textcomp}
\usepackage{epstopdf}
\usepackage{threeparttable}
\usepackage{subfigure}

\usepackage[english]{babel} 

\usepackage[hmarginratio=1:1,left=18mm,right=18mm,top=32mm,columnsep=20pt]{geometry} 
\usepackage[hang, small,labelfont=bf,up,textfont=it,up]{caption} 
\usepackage{booktabs} 

\usepackage{lettrine} 

\usepackage{enumitem} 
\setlist[itemize]{noitemsep} 

\usepackage{abstract} 

\usepackage{titlesec} 
\renewcommand\thesection{\Roman{section}} 
\renewcommand\thesubsection{\roman{subsection}} 
\titleformat{\section}[block]{\large\scshape\centering}{\thesection.}{1em}{} 
\titleformat{\subsection}[block]{\large}{\thesubsection.}{1em}{} 

\usepackage{fancyhdr} 
\usepackage{titling} 

\usepackage{hyperref} 

\pretitle{\begin{center}\Huge\bfseries} 
\posttitle{\end{center}} 
\title{DFTerNet: Towards 2-bit Dynamic Fusion Networks for Accurate Human Activity Recognition} 
\author{%
\textsc{Zhan Yang$^1$, Osolo Ian Raymond$^1$, ChengYuan Zhang$^1$, Ying Wan$^1$, Jun Long$^{1,2}$}\thanks{Corresponding author. This work was supported in part by the National Natural Science Foundation of China (61472450), the Natural Science Foundation of Hunan Province (2017JJ3417) and the Science and Technology Plan of Hunan (2016TP1003).} \\[1ex] 
\thanks{\copyright 2018 IEEE. Personal use of this material is permitted. Permission
from IEEE must be obtained for all other uses, in any current or future
media, including reprinting/republishing this material for advertising or
promotional purposes, creating new collective works, for resale or
redistribution to servers or lists, or reuse of any copyrighted
component of this work in other works. See \url{http://www.ieee.org/publications_standards/publications/rights/index.html} for more information}
\thanks{\textbf{IEEE ACCESS} DOI: 10.1109/ACCESS.2018.2873315}
\normalsize $^1$School of Information Science and Engineering, Central South University, Changsha 410083, China \\ 
\normalsize $^2$Network Resources Management and Trust Evaluation Key Laboratory of Hunan Province \\ 
\normalsize \href{mailto:junlong@csu.edu.cn}{junlong@csu.edu.cn} 
}
\date{\today} 
\renewcommand{\maketitlehookd}{%
\begin{abstract}
Deep Convolutional Neural Networks (DCNNs) are currently popular in human activity recognition (HAR) applications. However, in the face of modern artificial intelligence sensor-based games, many research achievements cannot be practically applied on portable devices (i.e., smart phone, VR/AR). DCNNs are typically resource-intensive and too large to be deployed on portable devices, thus this limits the practical application of complex activity detection. In addition, since portable devices do not possess high-performance Graphic Processing Units (GPUs), there is hardly any improvement in Action Game (ACT) experience. Besides, in order to deal with multi-sensor collaboration, all previous human activity recognition models typically treated the representations from different sensor signal sources equally. However, distinct types of activities should adopt different fusion strategies. In this paper, a novel scheme is proposed. This scheme is used to train 2-bit Convolutional Neural Networks with weights and activations constrained to $\text{\{-0.5,\ 0,\ 0.5\}}$. It takes into account the correlation between different sensor signal sources and the activity types. This model, which we refer to as DFTerNet, aims at producing a more reliable inference and better trade-offs for practical applications. It's known that quantization of weights and activations can substantially reduce memory size and use more efficient bitwise operations to replace floating or matrix operations to achieve much faster calculation and lower power consumption. Our basic idea is to exploit quantization of weights and activations directly in pre-trained filter banks and adopt dynamic fusion strategies for different activity types. Experiments demonstrate that by using a dynamic fusion strategy, it is possible to exceed the baseline model performance by up to $\sim$5\% on activity recognition datasets like the OPPORTUNITY and PAMAP2 datasets. Using the quantization method proposed, we were able to achieve performances closer to that of the full-precision counterpart. These results were also verified using the UniMiB-SHAR dataset. In addition, the proposed method can achieve $\sim$9$\times$ acceleration on CPUs and $\sim$11$\times$ memory saving. 
\end{abstract}
}

\begin{document}
\maketitle


\section{Introduction}
\label{sec:introduction}
Artificial Intelligence (AI), as an auxiliary technology in modern games, has played an indispensable role in improving gaming experience in the last decade. The film ``Ready Player One'' vividly shows the charm of future virtual games on the world. It demonstrates that one of the core technologies of virtual-realistic interaction is recognizing all kinds of complex activities.

Convolutional neural networks are very powerful and have been successfully used in many neural network models. They have been widely applied in lots of virtual-realistic interactive practical applications, e.g., object recognition~\cite{b1,b2,b3}, Internet of Things~\cite{b4,b5}, human activity recognition~\cite{b6,b7}. Its successes have been driven by the recent data explosion as well as the increase in model size.

\begin{figure}
  \centering
  \includegraphics[width=8cm]{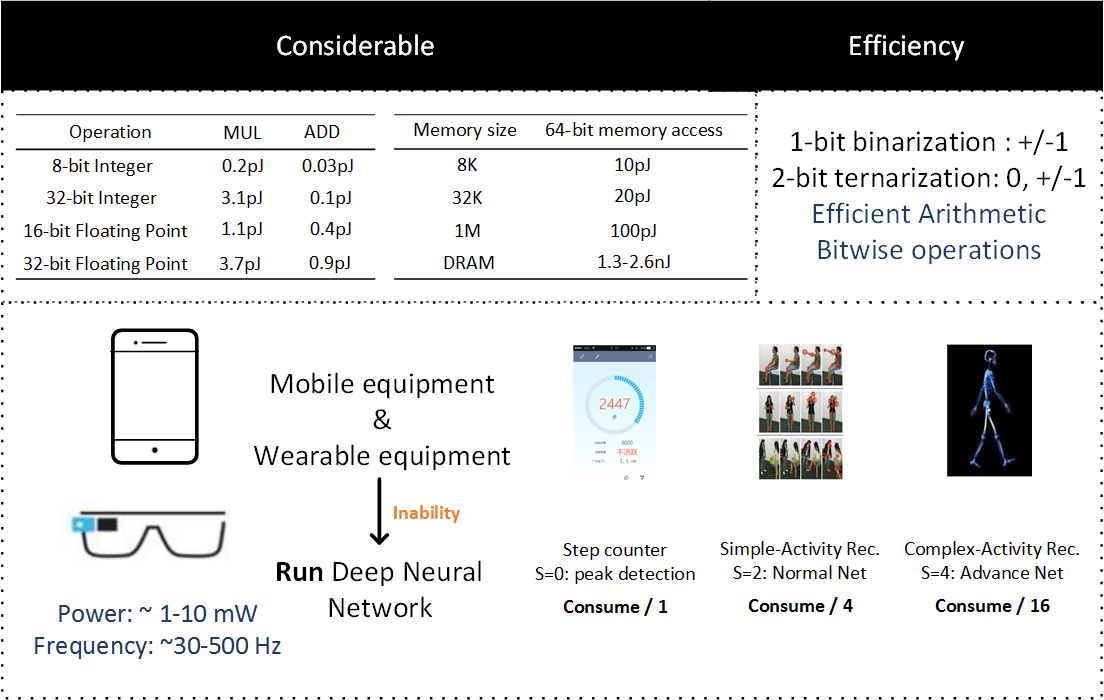}\\
  \caption{The rough numbers for the computations' energy consumption (45nm technology) ~\cite{b8} and some effective methods to deploy deep neural networks on portable devices.}\label{fig:F1}
\end{figure}

In spite of these successes, the large computational power and high memory requirements of these models restricts them from being deployed on portable devices that lack high-performance Graphical Processing Units (GPUs) as shown in Figure~\ref{fig:F1}. As far as we know, it is more interesting to operate sensor-based portable gaming devices which are able to recognize human activities. As a result of this, it is necessary to deploy advanced CNN, e.g., Inception-Nets~\cite{b9}, ResNets~\cite{b10} and VGG-Nets~\cite{b11} on smart portable devices. However, as the winner of the ILSVRC-2015 competition, ResNets-152~\cite{b10} is a classifier trained with nearly 19.4 million real-valued parameters, making it resource-intensive in different aspects. It is unable to run on portable devices for real-time applications, due to its high CPU/GPU workload and memory usage requirements. A similar phenomenon occurs in other state-of-the-art models, such as, VGG-Net~\cite{b11} and AlexNet~\cite{b12}.

Recently, in order to resolve the storage and computational problems~\cite{b13,b14} can be categorized into three methods, i.e., network pruning, low-rank decomposition and quantization of network. Among them, quantization of network has received more and more research focus. DCNNs with binary weights and activations have been designed ~\cite{b15,b16,b17}. Binary Convolutional Neural Networks (BCNNs) with weights and activations constrained to only two values (e.g., -1,+1), can bring great benefits to specialized machine learning hardware because of the following major reasons: (1) the quantized weights and activations reduce memory usage and model size by $\sim$32x compared to the full-precision version; (2) if networks are binary, then most multiply-accumulate operations (require hundreds of logic gates at least) can be replaced by popcount-XNOR operations (only require a single logic gate), which are especially well suited for FPGAs and ASICs~\cite{b18}.

However, the following problems limit the practical applications of above-mentioned idea: 1) quantization usually causes severe prediction accuracy degradation. The reported accuracy of obtained models is unsatisfactory on complex tasks (e.g., ImageNet dataset). More concretely, Rastegari \emph{et al.}~\cite{b17} shows that binary weights cause the accuracy of ResNet-18 to drop by about 9\% (GoogLenet drops by about 6\%) on the ImageNet dataset. It is obvious that there is a considerable gap in performance between the accuracy of a quantization model and the full-precision model. 2) in the case of practical applications, typical human activity recognition datasets containing multiple sensor data are collected from different positions on the body. This data can be fused at different stages within a convolutional neural network architecture~\cite{b19}. However, depending on the type of activity being performed, a sensor may contribute less or more to the overall result compared to other sensors depending on its location, and thus its fusion weight adjusted accordingly.

In light of these considerations, this paper proposes a novel quantization method and dynamic fusion strategy, to achieve deployments on an advanced high-precision and low-cost computation neural network model on portable devices. The main contributions of this paper are summarized as follows:

\begin{enumerate}
\item We propose a quantization function with an elastic scale parameter ($\boldsymbol{\epsilon}$-$\boldsymbol{Q}$) to quantize the entire full-precision convolutional neural network. The quantization of weights, activations and fusion weights are derived from the quantization function ($\boldsymbol{\epsilon }$-$\boldsymbol{Q}$) with different scale parameters ($\boldsymbol{\epsilon }$). We quantize the weights and activations to 2-bit values ${\{-0.5,\ 0,\ 0.5\}}$, and use a masked Hamming distance instead of a floating-point matrix multiplication. This setup is able to achieve an accuracy score close to that of the full-precision counterpart, using about $\sim$11$\times$ less memory and achieving a $\sim$9$\times$ speedup.

\item We introduce a dynamic fusion strategy for multi-sensors environmental activity recognition, which can decrease the computational complexity and improve the performance of model by reducing the representations from sensors that have ``less contribution'' during a particular activity. For sensors whose ``contribution'' (sub-network) is less than the others, we randomly reduce their representations through fusion weights, which are sampled from a Bernoulli distribution given by the scale parameter $\boldsymbol{\epsilon}$ from the quantization method. Experimental results show that by adopting a dynamic fusion strategy, we were able to achieve a higher accuracy level and lower memory usage than the baseline model.
\end{enumerate}

Ideally, using more quantization weights, activation modules and fusion strategy will result in better accuracy and eventually achieve a higher accuracy than the full-precision baselines. The training strategy can reduce the amount of computing power required and the energy consumption, thereby realizing the main objective of designing a system that can be deployed on portable devices. More importantly, adopting dynamic fusion strategies for different types of activities are more in line with the actual situation. This was verified by using both the quantization method and fusion strategy on the OPPORTUNITY and PAMAP2 datasets. Only the quantization method was applied on the UniMiB-SHAR dataset. \textbf{This is the first time that both quantization and dynamic fusion strategy were adopted for convolutional networks to achieve a high prediction accuracy on complex human activity recognition tasks.}

The remainder of this paper is structured as follows. In Section \ref{sec:Related_Work}, we briefly introduce the related works on human activity recognition, quantization models and methods for convolutional neural networks. In Section \ref{sec:Method}, we highlight the motivation of our method and provide some theoretical analysis for its implementation. In Section \ref{sec:EXPERIMENTS}, we introduce our experiment. Section \ref{sec:Result_and_Discussion} experimentally demonstrates the efficacy of our method and finally Section \ref{sec:Conclusion_and_Future_Work} states the conclusion and future work.

\section{Related Work}\label{sec:Related_Work}

\subsection{Convolutional Neural Networks for Human Activity Recognition}
Several advanced algorithms have been evaluated in the last few years on the human activity recognition. Hand-crafted features method~\cite{b20} uses simple statistical value (e.g., std, avg, mean, max, min, median, etc.) or frequency domain correlation features based on the signal Fourier transform to analyze the time series of human activity recognition data. Due to its simplicity to setup and low computational cost, it is still being used in some areas, but the accuracy cannot satisfy the requirement of modern AI games. Decision Trees~\cite{b21}, Support Vector Machines (SVM)~\cite{b22}, Random Forests~\cite{b19}, Dynamic Time Warping~\cite{b23} and Hidden Markov Models (HMMs)~\cite{b24,b25} for predicting action classes, work well when data are scarce and highly unbalanced. However, when faced with the activity recognition of complex high-level behaviors tasks, identifying the relevant features through these traditional approaches is time-consuming~\cite{b22}.

Recently, many researchers have adopted Convolutional Neural Networks (CNNs) to deploy human activity recognition system, such as ~\cite{b6,b22,b26,b27,b28}. Convolutional Neural Networks were based on the discovery of visual cortical cells and retain the spatial information of the data through receptive field. It is known that the power of CNNs stems in large part from their ability to exploit symmetries through a combination of weight sharing and translation equivariance. Also, with their ability to act as feature extractors, a plurality of convolution operators are stacked to create a hierarchy of progressive abstract features. Apart from image recognition ~\cite{b12,b29,b30}, NLP ~\cite{b31,b32} and video recognition ~\cite{b33}, more and more researches in recent years are using CNNs to learn sensor-based data representations for human activity recognition and have achieved remarkable performances ~\cite{b6,b34}. The model ~\cite{b28} consists of two or three temporal-convolution layers with ReLU activation function followed by max-pooling layer and a soft-max classifier, which can be applied over all sensors simultaneously. Yang \emph{et al.} ~\cite{b6} introduce a four temporal-convolutional layers on a single sensor, followed by a fully-connected layer and soft-max classifier and it shows that deeper networks can find correlations between different sensors.

Just like the works discussed above, we adopt convolutional neural networks to learn representations from wearable multi-sensor signal sources. However, these advanced high-precision models are difficult to deploy on portable devices, due to their computational complexity and energy consumption. As a result, quantization of convolutional neural networks has become a hot research topic. The aim is to reduce memory usage and computational complexity while maintaining an acceptable accuracy.

\subsection{Quantization Model of Convolutional Neural Networks}
The convolutional binary neural network is not a new topic. Inspired by neuroscience, the unit step function is used as an activation function in artificial neural networks ~\cite{b35}. The binary activation mode can use spiking response for computing and communication, which is an energy-efficient method because it consumes energy only when necessary ~\cite{b13}.

Recently, Binarized-neural-networks (BNNs) ~\cite{b16} have been used to quantize the weights and activations to binary values of each layer successfully. They proposed two binarization functions, the first is deterministic as shown in \eqref{eq:deterministic_method} and the second is stochastic as shown in \eqref{eq:stochastic_method}. Where $\tilde{w}$ is the binarized variable and $w$ is the full-precision variable, $\sigma( w )=clip( w+\frac{1}{2},\text{0,1})$ is the ``hard sigmoid'' function.
\begin{equation}\label{eq:deterministic_method}
\tilde{w}=\text{sign}(w) =\begin{cases}
	+\text{1\ }if\ w\geqslant \text{0;}\\
	-\text{1\ }if\ otherwise.\\
\end{cases}
\end{equation}
\begin{equation}\label{eq:stochastic_method}
\tilde{w}=\begin{cases}
	+\text{1\ }with\ probability\ p=\sigma(w) ;\\
	-\text{1\ }with\ probability\ 1-p.\\
\end{cases}
\end{equation}

Ternary Weight Network (TWN)~\cite{b36} constrains the weights to ternary values ($-1,0,1$) by referencing symmetric thresholds. In each layer, the quantization of TWN is shown in \eqref{eq:TWN}, where $\varDelta \approx 0.7\cdot E(|w|)\approx \frac{0.7}{n}\sum_{i=1}^{n}|w_i|$ is a positive threshold parameter. They claim a trade-off between model complexity and generalization.
\begin{equation}\label{eq:TWN}
\tilde{w}_i=f(w_i|\Delta) =\begin{cases}
	+\text{1,} & if\  w_i>\Delta ;\\
	\text{0,} & if\ |w_i|\leqslant \Delta ;\\
	-\text{1,} & if\ w_i<-\Delta .
\end{cases}
\end{equation}

DoReFa-Net ~\cite{b37} is derived from AlexNet, which has 1-bit weights, 2-bit activations and 6-bit gradients and it can achieve 46.1\% top-1 accuracy on ImageNet validation set. DoReFa-Net adopts a method as shown in \eqref{eq:DoRefa-net}, where $w$ and $\tilde{w}$ are the full-precision (original) and quantized weights, respectively, and $E(|w|)$ is the mean absolute value of weights.
 \begin{equation}\label{eq:DoRefa-net}
 \tilde{w}=E(|w|) \times \text{sign}(w).
 \end{equation}

\subsection{Quantization Method for Convolutional Neural Networks}
The idea of quantization of weights and activations was first proposed by ~\cite{b16}. The research showed the following two contributions: 1) the costly arithmetic operations between weights and activations in full-precision networks can be replaced with cheap bitcount and XNOR operations, which can result in significant speed improvements. Compared with the full-precision counterpart, 1-bit quantization reduces the memory usage by a factor of 32; and 2) in some visual classification tasks, using 1-bit quantization could achieve fairly good performance.

Some researchers ~\cite{b17,b38} have introduced easy, high-performance and accurate approximations to convolutional neural networks by quantizing the weights, and using a uniform quantization method, which first scales its value in the range $x\in[-\text{1,}\text{1}]$. Then it adopts the following $k$-bit quantization as shown in \eqref{eq:uniform_quantization_method}, where $round$ approximates continuous values to their nearest discrete states. The benefit of this quantization method is that when calculating the inner product of two quantized vectors, costly arithmetic calculations can be replaced by cheap operations. (e.g. bit shift, count operation) In addition, this quantization method is rule-based and thus easy to implement.
\begin{equation}\label{eq:uniform_quantization_method}
q_k\left( x \right) =2\left( \frac{round\left( \left( 2^k-1 \right) \left( \frac{x+1}{2} \right) \right)}{2^k-1}-\frac{1}{2} \right).
\end{equation}

Zhou \emph{et al.} ~\cite{b39} propose a network compression method called Incremental Network Quantization (INQ). After obtaining a network through training, the parameters (full-precision parameters) of each layer are first divided into two groups. The parameters in the first group are directly quantized and fixed. The other group of parameters through retraining compensated for the loss of accuracy caused by quantization. The above process iterates until all parameters are quantized. With incremental quantization, using weights with small-width values (e.g., 3-bit, 4-bit and 5-bit) results in almost no accuracy loss compared with the full-precision counterpart. The quantization method is shown in \eqref{eq:INQ}, where $w$ and $\tilde{w}$ are full-precision (original) and quantized weights, respectively, $lb$ and $ub$ are the lower and upper bounds of the quantized set, respectively.
\begin{equation}\label{eq:INQ}
\tilde{w}=\begin{cases}
	\text{sign} (w) \times 2^p\ if\ 3\times 2^{p-2}\leqslant |w|<3\times 2^{p-1};\\
	\ \ \ \ \ \ \ \ \ \ \ \ \ \ \ \ \ \ \ \ \ \ \ lb\leqslant p\leqslant ub;\\
	\text{sign} (w) \times 2^q\ if\ \left| w \right|\geqslant 2^{ub};\\
	\text{0}\ \ \ \ \ \ \ \ \ \ \ \ \ \ if\ |w|<2^{-lb-1}.\\
\end{cases}
\end{equation}

Wen \emph{et al}. ~\cite{b40} proposes a method as shown in \eqref{eq:TernGrad}, where $s_t:=\lVert g_t \rVert _{\infty}:=max(abs (g_t))$ is a scaler parameter, $\otimes$ is the Hadamard product, $abs(\cdot)$ respectively returns the $absolute$ value of each element. The method quantizes gradients to ternary values that can effectively improve clients-to-server communication in distributed learning.
\begin{equation}\label{eq:TernGrad}
\tilde{g}_t=s_t\cdot \mbox{sign} \left( g_t \right) \otimes b_t.
\end{equation}

Guo \emph{et al}. ~\cite{b41} propose greedy approximation, which instead tries to learn the quantization as shown in \eqref{eq:greedy approximation}, where $B_i$ is a binary filter, $\alpha_i$ are optimization parameters and input channels ($c$)$\times$ width ($w$)$\times$ height ($h$) is the size of the filter.
\begin{equation}\label{eq:greedy approximation}
\begin{cases}
	\begin{array}{l}
	\alpha _{i}^{*},B_{i}^{*}=\underset{\alpha _i,B_i}{arg\min}\lVert W-\sum_{i=1}^k{\alpha _iB_i}\rVert,\\
	\ \ \ \ \ \ \ s.t.\ \ \ \ \ \ B_i\in \left\{ -\text{1,}+1 \right\} ^{c\times w \times h}.\\
\end{array}\\
\end{cases}
\end{equation}

The greedy approximation expands to $k$-bit ($k>1$) quantization by minimizing the residue in order. Although not able to achieve a high-precision solution, the formulation of minimizing quantization error is very promising, and quantitative neural networks designed in this manner can be effectively deployed on modern portable devices.

\section{Method}\label{sec:Method}
In this section, we introduce our quantization method and dynamic fusion strategy, which is termed DFTerNet (Dynamic-Fusion-Ternary(2-bit)-Convolutional-Network) for convenience. We aim to recognize human activity extracted from IMU sensors. For this purpose, a fully-convolutional-based architecture is chosen and we focus on the recognition accuracy of the final model. During \textbf{train-time} (Training), we still use the full-precision network (the real-valued weights are retained and updated at each epoch). During \textbf{run-time} (Inference), we use ternary weights in convolution.

\subsection{Linear Mapping}
In this paper, we propose a quantization function $\boldsymbol{\epsilon}$-$\boldsymbol{Q}$ that converts a floating-point $x$ to its $k$-bitwidth signed integer. Formally, it can be defined as follows:
\begin{equation}\label{eq:quantization_function}
Q_k\left( x,\epsilon \right) =
\end{equation}
$$
Clip\left(\phi(k)\cdot round\left( \frac{x\cdot \boldsymbol{\epsilon }}{\phi \left( k \right)} \right) ,-1+\phi \left( k \right) ,1-\phi \left( k \right) \right),
$$

\noindent where $\phi \left( k \right) =2^{1-k},k>1\ \text{and}\ k\in \mathbb{N}_+$ is uniform distance, whose role is to perform a discretization of $k$-bit linear mapping of continuous and unbounded values, $\epsilon$ is a scale parameter, $round$ is the approximation function that approximates continuous values to their nearest discrete states, $clip$ function that clips unbounded values to [$-1+\phi(k)$,$1-\phi(k)$].

For example, when the scale parameter $\epsilon=1$, $Q_2(x,1)$ quantizes $\{-0.85,0.22,0.67\}$ to $\{-0.5,0,0.5\}$. Consider the scale parameter $\epsilon$, assume we set two different scale parameters: $\epsilon_1=1$ and $\epsilon_2=3$ corresponds to $Q_{k}^{1}\left( x,\epsilon _1 \right) =Clip\left( x\cdot 1,\ -0.5,\ 0.5\right)$ and $Q_{k}^{2}\left( x,\epsilon _2 \right) =Clip\left( x\cdot 3,\ -0.5,\ 0.5\right)$  . In that case $Q_{2}^{1}\left( 0.22,\ 1\right)$ is 0 and $Q_{2}^{2}\left(0.22,\ 3\right)$ is 0.5. Clearly, it can be seen that each quantization function can use the scale parameter to adjust the quantization threshold, clip differently to represent the input value.

\subsection{Approximate weights  $\boldsymbol{Q}_w\left( \cdot \right)$}
Consider that we use a $L$-layer CNN model. Suppose that learnable weights of each convolutional layer are represented as $W\in \mathbb{R}^{c_i\times c_o\times h\times w}$ , in which $c_i,c_o,w,h$ indicate the \textbf{i}nput-channel, \textbf{o}utput-channel, filter \textbf{w}idth and filter \textbf{h}eight, respectively. It is known that when using 32 bits (full-precision) floating-point arithmetic, storing all these weights would require a $32\times c_i\times c_o\times w\times h$ bit memory.

As stated above, at each layer, our goal is to estimate the real-weight filter $W\in \mathbb{R}^{c_i\times c_o\times h\times w}$ using 2-bit filter $T\in \left\{ \text{-0.5,\ 0,\ 0.5} \right\} ^{c_i\times c_o\times h\times w}$ . Generally, we define a reconstruction error $e$ as shown in \eqref{eq:reconstruction_error}:
\begin{equation}\label{eq:reconstruction_error}
e^2:=\lVert W-\alpha T \rVert ^2,
\end{equation}
where $\alpha$ describes a nonnegative scaling parameter. To retain the quantization network accuracy, the reconstruction error $e$ should be minimized. However, directly minimizing reconstruction error $e$ seems an NP-hard problem, so forcibly solving it will be very time-consuming ~\cite{b42}. In order to solve the above problem in a reasonable time, we need to find an optimal estimation algorithm, in which $T$ and $\alpha$ are sequentially learnt. That is to say, the goal is to solve the following optimization problem:
\begin{equation}\label{eq:optimization_problem}
\begin{aligned}
\underset{\alpha ,T}{\min}J\left( \alpha ,T \right) =\lVert W-\alpha T \rVert ^2,
\end{aligned}
\end{equation}
in which $T\in\text{\{-0.5,\ 0,\ 0.5\}}$, the $\lVert\cdot\rVert$ is defined as $\lVert W \rVert :=\left< W,W \right> ^{\text{1/}2}$ for any three-dimension tensor $W$.

One way to solve the optimization problem shown in \eqref{eq:optimization_problem} is to expand the cost function $J$ and take the derivate w.r.t. $\alpha$ and $T$, respectively. However, in this case, it must get correlation-dependence value of $\alpha$ and $T$. To overcome this problem, we use the quantization function $Q$ to quantize $W$ by \eqref{eq:quantization_function}:
\begin{equation}\label{eq:quantization_weight}
T=Q_w\left( W \right) =Q_{k_w}\left( W,\epsilon _w \right).
\end{equation}

In this work, we aim to quantize the real-weight filter $W$ to ternary values $\text{\{-0.5,0,0.5\}}$ , so the parameter $k_w=2$ and the threshold of weights are controlled by $\epsilon_w$ as shown in \eqref{eq:weight_boundary},
\begin{equation}\label{eq:weight_boundary}
\epsilon _w=\left( \frac{\xi}{n}\sum_{i=1}^n{\left| W \right|} \right) ^{-1},
\end{equation}
where $\xi$ is a shift threshold parameter which can be used to constrain thresholds.

With the ${T_i}'s$ fixed through \eqref{eq:quantization_weight}, Equation \eqref{eq:optimization_problem} becomes a linear regression problem:
\begin{equation}\label{eq:linear_regression_problem}
\underset{\alpha}{\min}J\left( \alpha\right) =\lVert W-\alpha T \rVert ^2,
\end{equation}

\noindent in which the ${T_i}'s$ serve as the bases in the matrix. Therefore, we can use the ``straight-through (ST) estimator'' ~\cite{b43} to back-propagate through $T_i$. This is shown in detail in Algorithm 1. Note that in \textbf{run-time}, only ($\clubsuit$) is required.

\begin{table}
\label{Algorithm 1}
\setlength{\tabcolsep}{3pt}
\begin{tabular}{p{240pt}}
\hline
\specialrule{0em}{2pt}{2pt}
\textbf{Algorithm 1} Training with ``straight-through (ST) estimator'' ~\cite{b43} on the forward and backward approach of an approximated convolution.\\
\specialrule{0em}{2pt}{2pt}
\hline
\specialrule{0em}{2pt}{2pt}
\textbf{Require} $k_w$-$bit$, shift parameter $\epsilon_w$. Assume $\mathcal{L}$ as the loss function, $\boldsymbol{I}$ and $\boldsymbol{O}$ as the input and output tensors of a convolutional layer respectively.\\
\specialrule{0em}{2pt}{2pt}
\textbf{A. Forward propagation:}\\
\specialrule{0em}{2pt}{2pt}
\ \ \ \ 1. $T=Q_w\left( W \right) =Q_{k_w}\left( W,\epsilon _w \right)$,\ \ \ \ \#Quantization\\
\ \ \ \ 2. Solve Eq. \eqref{eq:linear_regression_problem} for $\alpha$,\\
\ \ \ \ 3. $\boldsymbol{O}=\alpha Conv\left( T,I \right) $.\ \ \ \ \ \ \ \ \ \ \ \ \ \ \ \ \ \ \ \ ($\clubsuit$) \\
\specialrule{0em}{2pt}{2pt}
\textbf{B. Back propagation:}\\
\specialrule{0em}{1pt}{1pt}
\ \ \ \ By the chain rule of gradients and ST we have:\\
\specialrule{0em}{2pt}{2pt}
\ \ \ \ 1. $\frac{\partial \mathcal{L}}{\partial W}=\frac{\partial \mathcal{L}}{\partial O}\left( \alpha \frac{\partial O}{\partial T}\frac{\partial T}{\partial W} \right) \overset{\textbf{ST}}{\xlongequal{\quad}}\frac{\partial \mathcal{L}}{\partial O}\left( \alpha \frac{\partial O}{\partial T} \right) =\alpha \frac{\partial \mathcal{L}}{\partial T}$.\\
\specialrule{0em}{2pt}{2pt}
\hline
\end{tabular}
\end{table}

\subsection{Activation quantization $\boldsymbol{Q}_a\left( \cdot \right)$}
In order to avoid substantial memory consumption and computational requirement, which is caused by cumbersome floating-point calculations, we should use bitwise operation. Therefore, the activations as well as the weights must be quantized.

If activations are 1-bit values, we can quantize activations after they pass through a $clip$ function $\mathbb{C}$ similar to the activation quantization procedure in ~\cite{b37}. Formally, it can be defined as:
\begin{equation}
\mathbb{C}_a\left( x \right) =clip\left( x,-1,+1 \right).
\end{equation}

If activations are presented in $k_a$-$bit$, the quantized $A^q$ of real-value activations $A$ can be defined as:
\begin{equation}\label{eq:quantizaiton_activation}
A^q=Q_a\left( A \right) =Q_{k_a}\left( A,\epsilon _a \right).
\end{equation}

In this paper, we constrain the weights to ternary values $\text{\{-0.5,\ 0,\ 0.5\}}$. In order to transform the real-valued activation $a$ into ternary activation, we set the parameter $k_a=2$. The scale parameter $\epsilon_a$ controls the clip threshold and can be varied throughout the process of learning. Note that, quantization operations in networks will cause the variance of weights to be scaled compared to the original limit, which will cause exploding of network's outputs. XNOR-Net ~\cite{b17} proposes a filter-wise scaling factor calculated continuously with full precision to alleviate the amplification effect. In our experiment implementation, we control the activation threshold to attenuate the amplification effect by setting the scale parameter as
$$
\epsilon_a=min\{1,\frac{1}{2^{round(\tau)}}\},
$$
where $\epsilon_a$ is pre-defined constant for each layer, and $\epsilon_a$ will be updated by $\tau$ in each epoch:
$$
\tau =\frac{1}{m}\sum_{i=1}^m{\left| w_i \right|}/\underset{w_i,i=\text{1,..,}m}{\max}\left\{ w_i \right\},
$$
where $w$ is the trained weights of each layer. The forward and backward propagation of the activation are shown in detailed in Algorithm 2.

\begin{table}
\label{Algorithm 2}
\setlength{\tabcolsep}{3pt}
\begin{tabular}{p{240pt}}
\hline
\specialrule{0em}{2pt}{2pt}
\textbf{Algorithm 2} Training with ``straight-through (ST) estimator'' ~\cite{b43} on the forward and backward approach of the activation.\\
\specialrule{0em}{2pt}{2pt}
\hline
\specialrule{0em}{2pt}{2pt}
\textbf{Require} $k_a$-$bit$, shift parameter $\epsilon_a$, $1_{|A|\leqslant 0.5}$ can be seen as propagating the gradient through $hardtanh$, $\otimes$ indicates Hadamard product. Assume $\mathcal{L}$ as the loss function.\\
\specialrule{0em}{2pt}{2pt}
\textbf{A. Forward propagation:}\\
\specialrule{0em}{2pt}{2pt}
\ \ \ \ 1. $A^q=Q_{k_a}(A,\epsilon_a)$,\ \ \ \ \#Quantization\\
\specialrule{0em}{2pt}{2pt}
\textbf{B. Back propagation:}\\
\specialrule{0em}{2pt}{2pt}
\ \ \ \ 1. $\frac{\partial \mathcal{L}}{\partial A}=\frac{\partial \mathcal{L}}{\partial A^q}\otimes 1_{\left| A \right|\leqslant 0.5}$,\ \ \ \ \#using STE\\
\specialrule{0em}{2pt}{2pt}
\hline
\end{tabular}
\end{table}

\begin{figure*}
  \centering
  \includegraphics[width=17cm]{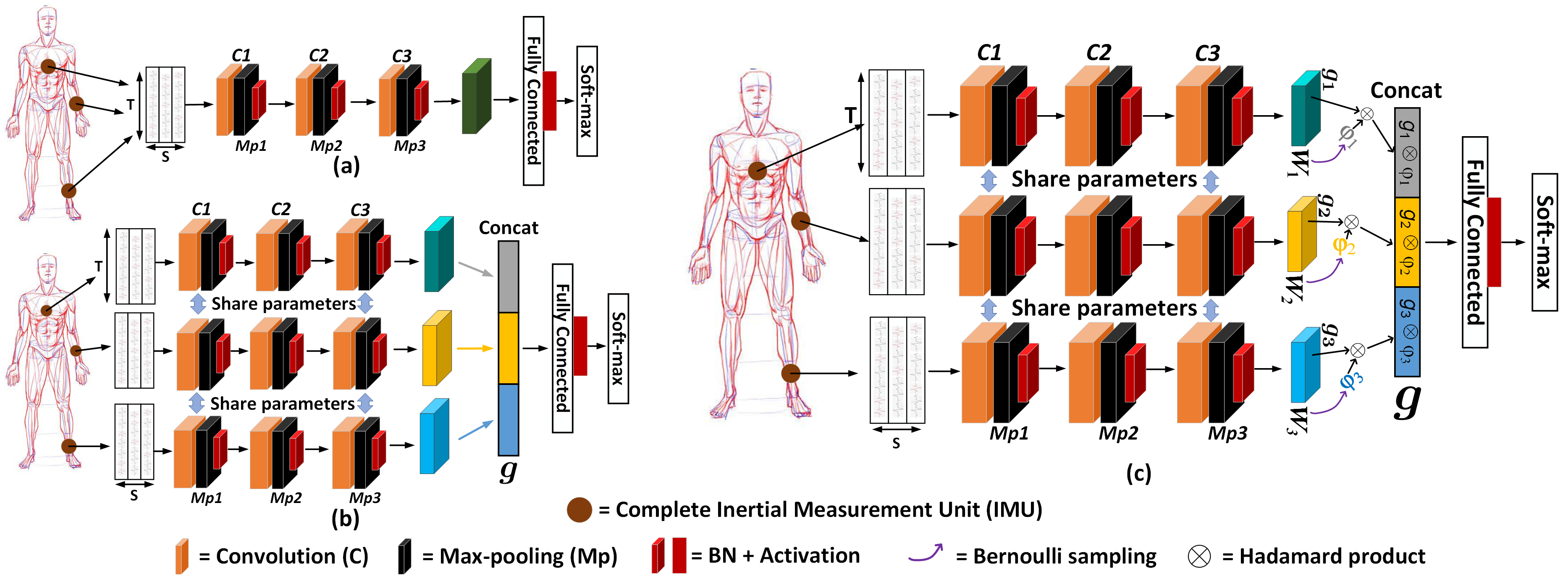}\\
  \caption{An overview of three fusion strategy methods and architecture of the hierarchical DFTerNet for activity recognition. (a) Early fusion, (b) Late fusion, (c) Dynamic fusion, are summarized in Section \ref{sec:fusion_stragegy}. From the left of each sub-fig, the multi-sensor signal sources from different positions are processed by a common convolutional network in (a) and three sub-convolutional networks in (b)\&(c). Input sensor signals of size $T$$\times$$S$, where $T$ denotes the length of features maps and $D$ the number of sensor channels. The \emph{C1}. \emph{C2}. \emph{C3}. ((kernel size), (siding stride), numbers of kernel) are ((11,1),(1,1),50), ((10,1),(1,1),40), ((6,1),(1,1),30), respectively. The \emph{Mp1}. \emph{Mp2}. \emph{Mp3}. size are (2,1), (3,1), (1,1), respectively. Neurons in \emph{fully connected layer} is 1000. The tensors $\varphi_1, \varphi_2, \varphi_3$ are the fusion weights.}\label{fig:F2}
\end{figure*}

\subsection{Scalability to Multiple Sensors (IMUs)}\label{sec:fusion_stragegy}
Each activity in the OPPORTUNITY and PAMAP2 datasets is collected by multi-sensors in different parts and each sensor is independent. For different types of activities, different sensors may not have the same ``contribution''. In order to improve the accuracy of our model, we conducted a comprehensive evaluation using different feature fusion strategies as shown in Figure \ref{fig:F2}. Note that the UniMiB-SHAR dataset only has 3-channels data (3D accelerometer), so we apply \textbf{early fusion}.

\textbf{Early fusion.} All joints from multi-sensors in different parts are stacked as input of the network ~\cite{b22,b44}.

\textbf{Late fusion.} Independent sensors in different signal sources through their own \emph{conv3} feature maps ($\boldsymbol{g}_{Conv3}\in \mathbb{R}^D$) are concatenated by fusion weights $\varphi_1=\varphi_2=\varphi_3=\{1\}^D$ as in ~\cite{b19,b45} and the feature maps $\boldsymbol{g}$ after fusion can be expressed as:
$$
\boldsymbol{g}=[\boldsymbol{g}_1,\boldsymbol{g}_2,\boldsymbol{g}_3].
$$

\textbf{Dynamic fusion.} Different parts of the body (different sensors locations) have different levels of participation in different types of activities. For example, for ankle-hand-based activities (e.g., running and jumping), the ``contribution'' of back-based sensor is lower than that of the sensors on the hands and ankles. In the case of hand-based activities (e.g., opening a drawer, closing a drawer), the ``contribution'' of the sensors in the ankles and back is lower than that of the hands, etc. Therefore, unlike in the \textbf{late fusion} method, the fusion weight settings of dynamic fusion are different. For notational simplicity, we refer to the last convolutional layer of CNNs, i.e., the $3^{rd}$ convolutional layer as \emph{conv3}, the full-precision weights and feature maps as $W_{conv3}:=W_p\in \mathbb{R}^m$ and $\boldsymbol{g}_p\in \mathbb{R}^D$ respectively, where $p\in{\{1,2,3\}}$ correspond to the fusion weights $\varphi_1, \varphi_2$, $\varphi_3$ $\in \mathbb{R}^D$. This is mainly because CNNs extract low-level features at the bottom layers and learn more abstract concepts such as the parts or more complicated texture patterns at the mid-level (i.e., \emph{conv3}). These mid-level representations are more informative compared to the higher-level representations~\cite{b46}. Hence, we propose a novel dynamic fusion method, which aims to \textbf{randomly} reduce the representations of less ``contribution'' signal sources. Dynamic fusion method can be considered as ``dynamic dropout method'', i.e., dynamic clip parameter by its weights (non-fixed parameter). Given a quantized weight $W_p^q=Q_{k}\left( W_p,\epsilon\right)$, each element of fusion weight $\varphi_{p_d}$ independently follows the Bernoulli distribution as shown in (24):
\begin{equation}\label{eq:Bernoulli_distribution}
\begin{cases}
	P\left( \varphi _{p_d}=1\left| W_p^q \right. \right) = \sum_{i=1}^{m}{|W_{p_i}^q|}/ m,\\
	P\left( \varphi _{p_d}=0\left| W_p^q \right. \right) =1-\sum_{i=1}^{m}{|W_{p_i}^q|}/ m,\\
\end{cases}
\end{equation}
where $\varphi_{p_d}$ and $W_{p_i}^q$ are the $d$-$th$ parameter of $\varphi_p$ and $W_p^q$ respectively.

\noindent\textbf{Train-time.} The full-precision weights $W_p$ are first quantized by \eqref{eq:quantization_function}:
\begin{equation}\label{eq:full-precision-dynamic_fusion_train-time_quantization_step_1}
W_p^q=Q_k\left( W_p,\epsilon\right).
\end{equation}

\begin{figure}
  \centering
  \includegraphics[width=8cm]{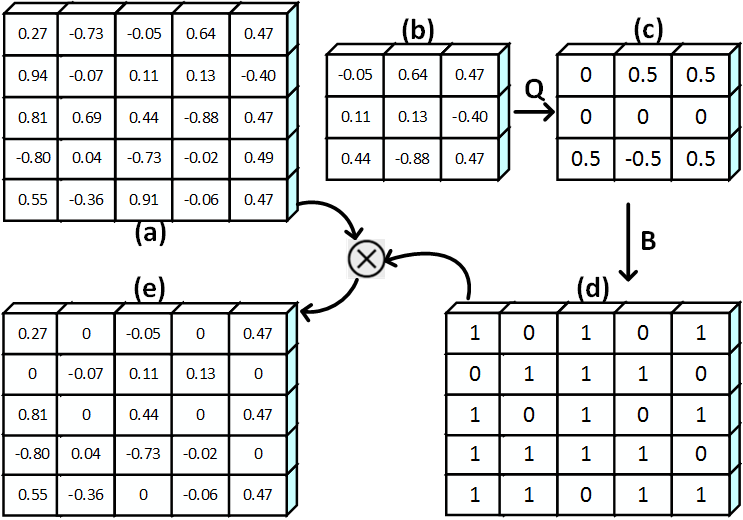}\\
  \caption{An example of the dynamic fusion processing when $k=2$ in \textbf{train-time}. (a) is the sub-network feature maps $\boldsymbol{g}_p$. (b) is the full-precision weights $W_p$. (c) represents the quantized weights $W_p^q$. (d) is the fusion weights $\varphi_p$. (e) is the feature maps $\boldsymbol{g}$ after fusion. $Q$ denotes the function $Q_2(w,\epsilon)$ from (b) to (c) is quantization function, which quantize the full-precision weights to 2-bit weights by using \eqref{eq:full-precision-dynamic_fusion_train-time_quantization_step_1}. The $B$ denotes $B\sim(W^q,m)$ from (c) to (d) is the Bernoulli distribution \eqref{eq:Bernoulli_distribution} that stochastically samples $\varphi_{p_d}$ to either 0 or 1, where $\otimes$ is Hadamard product.}\label{fig:F3}
\end{figure}

According to \eqref{eq:Bernoulli_distribution}, the generated fusion weight $\varphi_p$ as shown in \eqref{eq:full-precision-dynamic_fusion_train-time_quantization_step_2} is given by:
\begin{equation}\label{eq:full-precision-dynamic_fusion_train-time_quantization_step_2}
\varphi _p=\left[ \varphi _{p_1},\varphi _{p_2},...,\varphi _{p_D} \right].
\end{equation}
\textbf{Assumption.} The $Sub$-$network$-$(1,3)$ are the less ``contribution'' sub-networks. Thus, the fusion weight $\varphi _2$ is set to $\varphi _2=1^D$. The feature maps $\boldsymbol{g}$ after dynamic fusion strategy can be expressed as:
\begin{equation}\label{eq:full-precision-dynamic_fusion_train-time_quantization_step_3}
\boldsymbol{g}=concat\left[ \boldsymbol{g}_1\otimes \varphi _1, \boldsymbol{g}_2\otimes \varphi _2, \boldsymbol{g}_3\otimes \varphi _3 \right],
\end{equation}
where $\otimes$ denotes the Hadamard product. An example of this process is shown in Figure \ref{fig:F3}.

\noindent\textbf{Run-time.} The full-precision $W_p$ has been quantized in \textbf{run-time}. Therefore,~\eqref{eq:full-precision-dynamic_fusion_train-time_quantization_step_1} can be skipped and only \eqref{eq:Bernoulli_distribution},  \eqref{eq:full-precision-dynamic_fusion_train-time_quantization_step_2}, \eqref{eq:full-precision-dynamic_fusion_train-time_quantization_step_3} were used.

The use of a stochastic rounding method, instead of deterministic one, is chosen by TernGrad ~\cite{b40} and QSDG ~\cite{b47}. Some researchers (e.g. ~\cite{b16,b48}) have proven that stochastic rounding has an unbiased expectation and has been successfully on low-precision.

\subsection{Error and Complexity Analysis}\label{sec:proof_section}
\noindent\textbf{Reconstruction Error $\boldsymbol{e}$} According to \eqref{eq:reconstruction_error} and \eqref{eq:optimization_problem}, we have defined the reconstruction error $e$. In this section, we analyze the boundary that is satisfied by $e$.

\noindent\textbf{Theorem 1.} (Reconstruction Error Bound). The reconstruction error $e$ is bounded as
\begin{equation}\label{eq:reconstruction error boundary}
e^2\leqslant \lVert W_{\boldsymbol{\gamma}_{\epsilon}} \rVert ^2\left( \text{1}-\text{1}/\left| \boldsymbol{\gamma}_{\epsilon} \right| \right) ^k,
\end{equation}
where $\boldsymbol{\gamma }_{\epsilon}=\{ i \big| | w_i |>1/4\epsilon _w\}$ and $\left| \boldsymbol{\gamma }_{\epsilon} \right|$ denotes the number of elements in $\boldsymbol{\gamma }_{\epsilon}$.
\\
\\
\noindent\textbf{Proof.} We define $\mathring{W}_j$ which indicates the approximation residue after combining all the previously tensors as
\begin{equation}\label{eq:approximation_residue_tensor}
\mathring{W}_j=\begin{cases}
	W,\ if\ j=\text{0;}\\
	W-\sum_{i=0}^{j-1}{\alpha\ T_i,\ if\ j\geqslant 1.}\\
\end{cases}
\end{equation}

Through derivative calculations, \eqref{eq:reconstruction_error} is equivalent to
\begin{equation}\label{eq:reconstruction_error_through_derivative_calculations}
T_j=Q_{k_w}\left( W_j,\epsilon _w \right) \ \text{and}\ \ \alpha=\frac{2}{\left| \boldsymbol{\gamma }_{\epsilon} \right|}\sum_{j\in \boldsymbol{\gamma }_{\epsilon}}{\left| W_j \right|}.
\end{equation}

Since $j\in \boldsymbol{\gamma }_{\epsilon}$, we can obtain,
\begin{equation}\label{eq:proof_1}
\left< T_j,\mathring{W}_j \right> =\frac{1}{2} \sum{\left| \mathring{w}_j \right|\geqslant \frac{1}{2} \lVert \mathring{W}_j \rVert},
\end{equation}
in which $\mathring{w}_j$ is an entry of $\mathring{W}_j$. According to \eqref{eq:approximation_residue_tensor} and \eqref{eq:proof_1}, we have
\begin{equation}\label{eq:proof_2}
\begin{aligned}
\lVert \mathring{W}_{j+1} \rVert ^2&=\lVert \mathring{W}_j \rVert ^2-\alpha \left< T_j,\mathring{W}_j \right>
\\
&=\lVert \mathring{W}_j \rVert^2 \left( 1-\frac{2\sum_{j\in \boldsymbol{\gamma }_{\epsilon}}{\left| w_j \right|}\left< T_j,\mathring{W}_j \right>}{\left| \boldsymbol{\gamma }_{\epsilon} \right|\lVert \mathring{W}_j \rVert ^2} \right)
\\
&\leqslant \lVert \mathring{W}_j \rVert ^2\left( 1-\text{1/}\left| \boldsymbol{\gamma }_{\epsilon} \right| \right),
\end{aligned}
\end{equation}
in which $j$ varies from 0 to $k-1$. \ \ \ \ \ \ \ \ \ \ \ \ \ \ \ \ \ \ \ \ \ \ \ \ \ \ \ \ \ \ \ \ \ \ \ \ \ \ \ \ \ \ \ \ $\Box$

We can see from \textbf{Theorem 1} that, the reconstruction error $e^2$ follows an ``exponential decay'' with a rate $\text{1/}\left| \boldsymbol{\gamma }_{\epsilon} \right|$. It means that, given a small size $W$, i.e., $|\boldsymbol{\gamma }_{\epsilon}|$ is a small value, the reconstruction error $e$ algorithm can be quite good.

\noindent\textbf{Efficient Operations} Both modern CPUs and SoCs contain instructions to efficiently and massively compute 64-bit strings in short time cycles ~\cite{b49}. However, floating-point calculations require very complex logic. The calculation efficiency can be improved by several tens of times by adopting each bit-count operator instead of the 64-bit floating-point addition and multiplication calculation.

In the classic deep learning architecture, floating point multiplication is the most time-consuming part. However, when the weights and activations are ternary values, floating point calculations should be avoided. In order to efficiently reduce the computational complexity and time consumption, we have to design a new operation, which aims to replace the full-precision cumulant operation of input tensor $\mathbf{I}$ and filter $\mathbf{B}$. Some previous works ~\cite{b16,b17} on 1-bit networks have been successfully implemented using Hamming space calculation\footnote{The Hamming space can be used to calculate matrix multiplication and its inner-products.} (bit-counting) as a replacement for matrix multiplication. For example, $\mathbf{I,B}\in \text{\{0,1\}}^m$, the matrix multiplication can be replaced by \eqref{eq:basic_hamming_space}:
\begin{equation}\label{eq:basic_hamming_space}
IB=m-\varXi \left\{ I_i\oplus B_i \right\},
\end{equation}
where $\varXi$ defines a bit-count over the $m$ bits in the rows of $\mathbf{I}$ and $\mathbf{B}$, and $\oplus$ an exclusive OR operator.

In this paper, we aim to extend the concept to $k$-bit networks. The quantized input tensor $\mathbf{I}$ and filter $\mathbf{W}$ can be denoted as $\mathbf{I^q}=Q_{k}(\mathbf{I},\epsilon_i) \in \mathbb{R}^m$, $\mathbf{W^q}=Q_{k}(\mathbf{W},\epsilon_w) \in \mathbb{R}^m$, where the value of $\mathbf{I^q}$ and $\mathbf{W^q}$ are composed of $sign$, $positive$-$integer\ numerator(pin)$ and $positive$-$integer\ denominator(pid)$. Given a fixed $k$ value, the $pid$ is fixed as well. Therefore, we define two tensors as $\mathbf{I^s} \in \{0,1\}^m,\mathbf{I^v} \in \{0,1\}^{(k-1)\times m}$ and $\mathbf{W^s} \in \{0,1\}^m,\mathbf{W^v} \in \{0,1\}^{(k-1)\times m}$ to store $sign$ and $pin$, respectively. (Note that the superscript $s$ and $v$ mean $sign$ and $value$ respectively).
The values of $\boldsymbol{I}_{i}^{\boldsymbol{s}},\boldsymbol{W}_{i}^{\boldsymbol{s}}$ and $\boldsymbol{I}_{i}^{\boldsymbol{v}},\boldsymbol{W}_{i}^{\boldsymbol{v}}$ as
$$
\boldsymbol{I}_{i}^{\boldsymbol{s}},\boldsymbol{W}_{i}^{\boldsymbol{s}}=\begin{cases}
	\text{1\ ,\ }if\ \boldsymbol{I}_{i}^{\boldsymbol{q}},\boldsymbol{W}_{i}^{\boldsymbol{q}}\ is\ positive;\\
	\text{0,\ }else.\\
\end{cases}
$$
$$
\boldsymbol{I}_{i}^{\boldsymbol{v}},\boldsymbol{W}_{i}^{\boldsymbol{v}}=\begin{cases}
	pin\ to\ binary\ ,\ if\ \boldsymbol{I}_{i}^{\boldsymbol{q}},\boldsymbol{W}_{i}^{\boldsymbol{q}}\ne \text{0;}\\
	\text{0,\ }else.\\
\end{cases}
$$

In this work, our goal is to replace matrix multiplication with the notion of bit-counting in $k$-bit convolutional networks. Imagine that $k$=$2$, $\mathbf{I^q},\mathbf{W^q} \in \{-0.5,0,0.5\}^m$. Therefore, the inner-product calculation can be used with two bit-counts in Hamming space:
\begin{equation}\label{eq:our_Hamming_space}
\begin{aligned}
\boldsymbol{I}^{\boldsymbol{q}}\boldsymbol{W}^{\boldsymbol{q}}&=(\varXi \{\left( \overline{\boldsymbol{I}_{i}^{\boldsymbol{s}}\oplus \boldsymbol{W}_{i}^{\boldsymbol{s}}} \right) \&\left( \boldsymbol{I}_{i}^{\boldsymbol{v}}\&\boldsymbol{W}_{i}^{\boldsymbol{v}} \right)\}\\
&-\varXi \{\left( \boldsymbol{I}_{i}^{\boldsymbol{s}}\oplus \boldsymbol{W}_{i}^{\boldsymbol{s}} \right) \&\left( \boldsymbol{I}_{i}^{\boldsymbol{v}}\&\boldsymbol{W}_{i}^{\boldsymbol{v}} \right)\}) \cdot \phi(k)^2,
\end{aligned}
\end{equation}
where $\overline{A\oplus B}$ defines the negated XOR, $\&$ an AND operator.
Note that, if $k\geqslant3$, the behavior of the element-wise operator must be custom.

\noindent\textbf{Batch Normalization} In previous works, weights are quantized to binary values by using a $sign$ function ~\cite{b16} and to ternary values by using a positive threshold parameter $\Delta$ ~\cite{b36} during \textbf{train-time}. However, neural networks with quantized weights all failed to converge without batch normalization, because the quantized values are rather discretization for full-precision values. Batch normalization ~\cite{b50} efficiently avoids the exploding and vanishing gradients problem. In this part, we briefly discuss the batch normalization operation which might increase extra computational cost. Simply, batch normalization is an affine function:
\begin{equation}\label{eq:BN affine function}
BN\left(x \right) =\gamma \cdot \hat{x}+\beta,
\end{equation}
where $\hat{x}=\frac{x-\mu}{\sigma}$, $\mu$ and $\sigma$ are the mean and standard deviations respectively, $\gamma$ and $\beta$ are scale and shift parameters respectively. More specifically, a batch normalization can be quantized in 2-bit values by the following quantization method:
\begin{equation}\label{eq:BN_quantization_method}
Q_{k_b}\left( BN\left( x \right) ,\epsilon \right) =
\end{equation}
\begin{small}
$$
clip\left( \phi \left( k_B \right) \cdot round\left( \frac{BN\left( x \right) \cdot \epsilon}{\phi \left( k_B \right)} \right) ,-1+\phi \left( k_B \right) ,1-\phi \left( k_B \right) \right),
$$
\end{small}

\noindent where $k_b$=$2$. Equation \eqref{eq:BN_quantization_method} can be converted to the following:
\begin{equation}\label{eq:BN_convert}
\begin{split}
Q_2\left( BN\left( x \right) \right) =\begin{cases}
	+\text{0.5,\,\,}2\left( \gamma \cdot \hat{x}+\beta \right) \cdot \epsilon >\frac{1}{2};\\
	\text{0,\,\,}\left| 2\left( \gamma \cdot \hat{x}+\beta \right) \cdot \epsilon \right|\leqslant \frac{1}{2};\\
	-\text{0.5,\,\,}else.\\
\end{cases}
\\
=\begin{cases}
	+\text{0.5,\ }\hat{x}> \left( \frac{1}{4\epsilon}-\beta \right) /\gamma ;\\
	\text{0,\ }-\left( \frac{1}{4\epsilon}+\beta \right) /\gamma \leqslant \hat{x} \leqslant \left( \frac{1}{4\epsilon}-\beta \right) /\gamma ;\\
	-\text{0.5,\ }else.\\
\end{cases}
\end{split}
\end{equation}
Therefore, batch normalization will be accomplished at no extra cost.

\section{EXPERIMENTS}\label{sec:EXPERIMENTS}
To demonstrate the usefulness of quantization methods and fusion strategies on convolutional neural networks for high-precision human activity recognition on portable devices, we explore the activity recognition on three well-known datasets. The extension of our quantization methods and fusion strategies to activity recognition is straightforward. Providing better game experience for virtual-realistic interactive games on VR/AR devices and portable devices. Therefore, the memory requirements and the quantized weights of each layer are also analyzed in detail. Many natural activities are complex, involve several parts of the body and often very faint or subtle making recognition very difficult. Therefore, networks with better generalization ability to robustly fuse the data features of different parts of sensor are necessary, at the same time, an automatic method should depict the sketch of the activity feature and accurately recognize the activity.

The primary parameter of any experimental setup is the choice of datasets. To choose the optimal datasets for this study, we considered the complexity and richness of the datasets. Based on the background of our research, we selected the OPPORTUNITY ~\cite{b51}, PAMAP2 ~\cite{b52} and UniMiB-SHAR ~\cite{b53} benchmark datasets for our experiments.

\subsection{Data Description and Performance Measure}
\subsubsection{OPPORTUNITY}
The OPPORTUNITY public dataset has been used in many open activity recognition challenges. It contains four subjects performing 17 different (morning) Activities of Daily Life (ADLs) in a sensor-rich environment, as listed in Table \ref{tab:OPPORTUNITY DATASET}. They were acquired at a sampling frequency of 30Hz equipping 7 wireless body-worn inertial measurement units (IMUs). Each IMU consists of a 3D accelerometer, 3D gyroscope and a 3D magnetic sensor, as well as 12 additional 3D accelerometers placed on the back, arms, ankles and hips, accounting for a total of 145 different sensor channels. During the data collection process, each subject performed a session 5 times with ADL and 1 drill session. During each ADL session, subjects were asked to perform the activities naturally-named ``ADL1'' to ``ADL5''. During the drill sessions, subjects performed 20 repetitions of each of the 17 ADLs of the dataset. The dataset contains about 6 hours of information in total, and the data are labeled on a timestamp level. In our experiment, the training and testing sets have \textbf{63}-\textbf{D}imensions (\textbf{36}-\textbf{D} on hand, \textbf{9}-\textbf{D} on back and \textbf{18}-\textbf{D} on ankle, respectively).

In this paper, the models were trained on the data of ADL1, ADL2, ADL3, drill session, and the model test on the data of ADL4, ADL5.

\subsubsection{PAMAP2}
The PAMAP2 dataset contains recordings from 9 subjects who participated in carrying out 12 activities, including household activities and a variety of exercise activities as shown in Table \ref{tab:PAMAP2_DATASET}. The IMU and HR-monitor are attached on the hand, chest and ankle and they are sampled at a constant sampling rate of 100Hz\footnote{Note that, following ~\cite{b26}, the PAMAP2 dataset is downsampled to $\frac{100\text{Hz}}{3}=33.33$Hz, in order to have a temporal resolution comparable to the OPPORTUNITY dataset.}. The accelerometer, gyroscope, magnetometer, temperature and heart rate contain 40 sensors and are recorded from IMU over 10 hours in total. In our experiments, the training and testing sets have \textbf{36}-\textbf{D}imensions (\textbf{12}-\textbf{D} on hand, \textbf{12}-\textbf{D} on back and \textbf{12}-\textbf{D} on ankle, respectively).

In this paper, data from subjects 5 and 6 are used as testing sets and the remaining data are used for training.

\subsubsection{UniMiB-SHAR}
The UniMiB-SHAR dataset collected data from 30 healthy subjects (6 male and 24 female) acquired using the 3D-accelerometer of a Samsung Galaxy Nexus I9250 with Android OS version 5.1.1. The data are sampled at a constant sampling rate of 50 Hz, and split into 17 different activity classes, 9 safety activities and 8 dangerous activities (e.g., a falling action) as shown in Table \ref{tab:UniMiB-SHAR_DATASET}. Unlike the OPPORTUNITY dataset, the dataset does not have any NULL class and remains relatively balanced. In our experiments, the training and testing sets have \textbf{3}-\textbf{D}imensions.

\subsubsection{Performance Measure}
ADL datasets like the OPPORTUNITY dataset are often highly unbalanced. For this dataset, the overall classification accuracy is not an appropriate measure of performance, because the activity recognition rate of the majority classes might skew the performance statistics to the detriment of the least represented classes. As a result, many previous researches such as ~\cite{b22} show the use of an evaluation metric independent of the class repartition---$F1$-score. The $F1$-score combines two measures: the \textbf{precision} $p$ and the \textbf{recall} $r$: $p$ is the number of correct positive examples divided by the number of all positive examples returned by the classifier, and $r$ is the number of correct positive results divided by the number of all positive samples. The $F1$-score is the harmonic average of $p$ and $r$, where the best value is at 1 and worst at 0. In this paper, we use an additional evaluation metric to make the comparison with them easier: the weighted $F1$-Score (Sum of class $F1$-scores, weighted by the class proportion):
 \begin{equation}\label{eq:weight_F1-score}
F_w=2\sum_G{w_g\frac{p_g\cdot r_g}{p_g+r_g}},
 \end{equation}
where $w_g=N_g/N_{total}$ and $N_g$ is the number of samples in class $g$, and $N_{total}$ is the total number of samples.

\begin{table*}
\centering
\caption{Classes and proportions of the OPPORTUNITY dataset}
\setlength{\tabcolsep}{3pt}
\begin{tabular}{cccc}
\hline
\specialrule{0em}{2pt}{2pt}
Class&
Proportion&
Class&
Proportion\\
\specialrule{0em}{2pt}{2pt}
\hline
\specialrule{0em}{1pt}{1pt}
Open Door 1/2&
1.87\%/1.26\%&
Open Fridge&
1.60\%\\
Close Door 1/2&
6.15\%/1.54\%&
Close Fridge&
0.79\%\\
Open Dishwasher&
1.85\%&
Close Dishwasher&
1.32\%\\
Open Drawer 1/2/3&
1.09\%/1.64\%/0.94\%&
Clean Table&
1.23\%\\
Close Drawer 1/2/3&
0.87\%1.69\%/2.04\%&
Drink from Cup&
1.07\%\\
Toggle Switch&
0.78\%&
NULL&
72.28\%\\
\specialrule{0em}{1pt}{1pt}
\hline
\end{tabular}
\label{tab:OPPORTUNITY DATASET}
\end{table*}

\begin{table*}
\centering
\caption{Classes and proportions of the PAMAP2 dataset}
\setlength{\tabcolsep}{3pt}
\begin{tabular}{cccc}
\hline
\specialrule{0em}{2pt}{2pt}
Class&
Proportion&
Class&
Proportion\\
\specialrule{0em}{2pt}{2pt}
\hline
\specialrule{0em}{1pt}{1pt}
Lying&
6.00\%&
Sitting&
5.78\%\\
Standing&
5.92\%&
Walking&
7.45\%\\
Running&
3.06\%&
Cycling&
5.13\%\\
Nordic walking&
5.87\%&
Ascending stairs&
3.66\%\\
Descending stairs&
3.27\%&
Vacuum cleaning&
5.47\%\\
Ironing&
7.44\%&
House cleaning&
5.84\%\\
Null&
35.12\%\\
\specialrule{0em}{1pt}{1pt}
\hline
\end{tabular}
\label{tab:PAMAP2_DATASET}
\end{table*}

\begin{table*}
\centering
\caption{Classes and proportions of the UniMiB-SHAR dataset}
\setlength{\tabcolsep}{3pt}
\begin{tabular}{cccc}
\hline
\specialrule{0em}{2pt}{2pt}
Class&
Proportion&
Class&
Proportion\\
\specialrule{0em}{2pt}{2pt}
\hline
\specialrule{0em}{1pt}{1pt}
StandingUpfromSitting&
1.30\%&
Walking&
14.77\%\\
StandingUpfromLaying&
1.83\%&
Running&
16.86\%\\
LyingDownfromStanding&
2.51\%&
Going Up&
7.82\%\\
Jumping&
6.34\%&
Going Down&
11.25\%\\
F(alling) Forward&
4.49\%&
F and Hitting Obstacle&
5.62\%\\
F Backward&
4.47\%&
Syncope&
4.36\%\\
F Right&
4.34\%&
F with ProStrategies&
4.11\%\\
F Backward SittingChair&
3.69\%&
F Left&
4.54\%\\
Sitting Down&
1.70\%&
&
\\
\specialrule{0em}{1pt}{1pt}
\hline
\end{tabular}
\label{tab:UniMiB-SHAR_DATASET}
\end{table*}

\subsection{Experimental Setup}\label{sec:Experimental_Setup}
\noindent\textbf{Sliding Window} Our selected data are recorded continuously. We can think of the continuous-HAR data feature as a video feature. We use a sliding time-window of fixed length to segment the data. Each segmented data can be viewed as a frame in the video (a picture). We define $T$, $S$ and $\sigma$ as the length of the time-window, the number of sensor channels and the sliding stride, respectively. Through the above segment approach, each ``picture'' consists of a $T$$\times$$S$ matrix. We set the segment parameters like ~\cite{b44}, use a time-window of 2s on the OPPORTUNITY and PAMAP2 datasets, resulting in $T$=64, and $\sigma=3$. On the UniMiB-SHAR dataset, a time-window of 2s was used, resulting in $T$=96. Due to the timestamp-level labeling, each segmented data can usually contain multiple labels. A majority labeling that appears most frequently is chosen from among those of $T$ timestamps.

\noindent\textbf{Dynamic Fusion Weights} Our selected datasets (OPPORTUNITY and PAMAP2) include two families of human activity recognition samples: that of \textbf{p}eriodic activities (the locomotion category of the OPPORTUNITY dataset and the whole of PAMAP2 dataset) and that of \textbf{s}poradic activities (the gestures category of the OPPORTUNITY dataset). For designing the dynamic fusion strategies of the two families of activities, we design two groups of feature maps after dynamic fusion strategies. In \textbf{p}eriodic activities ($\boldsymbol{g}_p$), we take into account the fact that back-based sensors have less ``contribution''. Thus, we set the fusion weights $\varphi_2=\varphi_3=1^D$. In \textbf{s}poradic activities ($\boldsymbol{g}_s$), we consider the fact that both back-based and ankle-based sensors have less ``contribution''. Accordingly, we set the fusion weight $\varphi_2=1^D$. Formally, in \textbf{train-time} and \textbf{run-time}, according to \eqref{eq:Bernoulli_distribution}, \eqref{eq:full-precision-dynamic_fusion_train-time_quantization_step_1} and \eqref{eq:full-precision-dynamic_fusion_train-time_quantization_step_2}, the feature maps $\boldsymbol{g}_p$ and $\boldsymbol{g}_s$ after dynamic fusion strategies can be expressed as~\eqref{eq:Dynamic_fusion_parameters_training_time} and the effectiveness of these strategies we designed for the two families of human activity recognition is verified in Section~\ref{subsec:MSF}.
\begin{equation}\label{eq:Dynamic_fusion_parameters_training_time}
\begin{split}
\boldsymbol{g}_p=concat\left[ \boldsymbol{g}_1\otimes \varphi_1,\boldsymbol{g}_2\otimes \varphi_2, \boldsymbol{g}_3\otimes \varphi_3 \right],
\\
\boldsymbol{g}_s=concat\left[ \boldsymbol{g}_1\otimes \varphi_1,\boldsymbol{g}_2\otimes \varphi_2, \boldsymbol{g}_3\otimes \varphi_3 \right].
\end{split}
\end{equation}

\noindent\textbf{Pooling Layer} The role of common pooling layers is to find the maximum (max-pooling) or the average (avg-pooling) of output of each filter. Our experiments do not use the avg-pooling because the average operation will generate other values except $\{-0.5,\ 0,\ 0.5\}$. However, we observe that using max-pooling on $\{-0.5,\ 0,\ 0.5\}$ will increase the probability distribution of $\{0.5\}$, resulting in a noticeable drop in recognition accuracy. Therefore, we put the max-pooling layer before the batch normalization (BN) and activation (A) layers.

\subsection{Baseline Model}
The aim of this paper is not necessarily to exceed current state-of-the-art accuracies, but rather demonstrates and analyzes the impact of network model quantization and fusion strategy. Therefore, the benchmark model we used should not be very complex, because increasing the network topology and computational complexity to improve model performance runs counter to the aim of deploying advanced networks in portable devices. In this paper, we considered improving the performance of the model through a training strategy that is more in line with the practical applications. We therefore chose a CNN architecture ~\cite{b44} as the baseline model. It contains three convolutional blocks, a dense layer and a soft-max layer. Each convolutional kernel performs a 1D convolutional layer on each sensor channel independently over the time dimension. To fairly evaluate the calculation consumption and memory usage of binarization (1-bit) and ternarization (2-bit) on the CNNs, we employ the same number of channels and convolution filters for all comparison models. For early fusion Binary Convolutional Networks~\cite{b16} (termed BNN) and late fusion Binary Convolutional Networks~\cite{b16} (termed FBNN)\footnote{Note that, when the value of quantized weights is $\{-1,+1\}$, each element of fusion weights $\varphi$ is all 1. Therefore, dynamic fusion Binary Convolutional Networks can be considered as FBNN.}, see~\eqref{eq:basic_hamming_space}.

Layer-wise details are shown in Table \ref{tab:computational}, in which ``Conv2'' is the most computationally expensive and ``Fc'' commits the most memory. For example, using floating-point precision on OPPORTUNITY dataset, the entire model requires approximately 82MFLOPs\footnote{Note that FLOPs consist of the same number of FMULs and FADDs.} and approximately 2 million weights, thus 0.38MBytes of storage for the model weights. During \textbf{train-time} the model requires more than 12GBytes of memory (batch size of 1024), for inference during \textbf{run-time} this can be reduced to approximately 1.8GBytes (2-bit).

\begin{table*}[tp]
\centering
\begin{threeparttable}
  \centering
  \caption{Details of the learnable layers in our experimental model.}
  \label{tab:computational}
    \begin{tabular}{ccccc}
    \toprule
    &\multicolumn{2}{c}{OPPORTUNITY}&\multicolumn{2}{c}{PAMAP2}\cr
    \midrule
    \specialrule{0em}{2pt}{2pt}
    Layer Name&Params (b)&FLOPs&Params (b)&FLOPs\cr
    \specialrule{0em}{2pt}{2pt}
    \hline
    \specialrule{0em}{2pt}{2pt}
    Conv1&$\sim$0.6k&$\sim$4.84M&$\sim$0.6k&$\sim$2.76M\cr
    \specialrule{0em}{0.5pt}{0.5pt}
    \specialrule{0em}{0.5pt}{0.5pt}
    Conv2&$\sim$20k&$\sim$68.18M&$\sim$20k&$\sim$38.96M\cr
    \specialrule{0em}{0.5pt}{0.5pt}
    \specialrule{0em}{0.5pt}{0.5pt}
    Conv3&$\sim$7.2k&$\sim$5.47M&$\sim$7.2k&$\sim$3.12M\cr
    \specialrule{0em}{0.5pt}{0.5pt}
    \specialrule{0em}{0.5pt}{0.5pt}
    Fc&$\sim$1.89M&$\sim$3.78M&$\sim$1.89M&$\sim$2.16M\cr
    \specialrule{0em}{2pt}{2pt}
    \hline
    \hline
    \specialrule{0em}{2pt}{2pt}
    &\multicolumn{2}{c}{UniMiB-SAHR}\cr
    \specialrule{0em}{2pt}{2pt}
    \hline
    \specialrule{0em}{2pt}{2pt}
    Layer Name&Params (b)&FLOPs\cr
    \specialrule{0em}{2pt}{2pt}
    \hline
    \specialrule{0em}{2pt}{2pt}
    Conv1&$\sim$0.6k&$\sim$0.23M\cr
    \specialrule{0em}{0.5pt}{0.5pt}
    \specialrule{0em}{0.5pt}{0.5pt}
    Conv2&$\sim$20k&$\sim$3.25M\cr
    \specialrule{0em}{0.5pt}{0.5pt}
    \specialrule{0em}{0.5pt}{0.5pt}
    Conv3&$\sim$7.2k&$\sim$0.26M\cr
    \specialrule{0em}{0.5pt}{0.5pt}
    \specialrule{0em}{0.5pt}{0.5pt}
    Fc&$\sim$1.89M&$\sim$0.18M\cr
    \bottomrule
    \end{tabular}
\end{threeparttable}
\end{table*}

\subsection{Implementation Details}\label{sec:Implementation_Details}
In this section, we provide the implementation details of the architecture of the convolution neural network. Our method is implemented with Pytorch. The model is trained with mini-batch size of 1024, the activation scale parameter $\epsilon_a$ is initialized to 1, 50 epochs and using the AdaDelta with default initial learning rate ~\cite{b54}. The cross-entropy was used as the loss function. A soft-max function is used to normalize the output of the model. The probability $P$ that a sequence $\boldsymbol{S}$ belongs to the $i$-$th$ class is given by \eqref{eq:softmax_function}:
\begin{equation}\label{eq:softmax_function}
P\left( G_i\left| \boldsymbol{S} \right. \right) =\frac{e^{o_i}}{\sum_{j=1}^G{e^{o_j}}},\ i=\text{1,2,3,...,}G,
\end{equation}
where $o=(o_1,o_2,o_3,...,o_G)^T$ is the output of the model, $G$ is the number of activities.

Experiments were carried out on a platform with an Intel 2$\times$ Intel E5-2600 CPU, 128G RAM and a NVIDIA TITAN Xp 12G GPU. The hyper-parameters of the model are provided in Figure \ref{fig:F2}\footnote{The early fusion is common convolutional neural network architecture (can be regarded as a sub-network). Therefore, the hyper-parameters of early fusion is equal to any sub-network of late fusion or dynamic fusion.}. The training procedure, i.e., DFTerNet, is summarized in Algorithm 3.

\begin{table}
\setlength{\tabcolsep}{3pt}
\begin{tabular}{p{240pt}}
\hline
\specialrule{0em}{2pt}{2pt}
\textbf{Algorithm 3} Training a $L$-layer DFTerNet, $\mathcal{L}$ is the loss function for minibatch, $1_{|a|\leqslant 0.5}$ can be seen as propagating the gradient through $hardtanh$ and $\lambda$ is the learning rate decay factor. $\otimes$ indicates Hadamard product. BatchNorm() specifies how to batch-normalize the output of convolution. BackBatchNorm() specifies how to backpropagate through the normalization ~\cite{b50}. Update() specifies how to update the parameters when their gradients are known, such as AdaDelta ~\cite{b54}.\\
\specialrule{0em}{2pt}{2pt}
\hline
\specialrule{0em}{2pt}{2pt}
\textbf{Require} A minibatch of inputs and targets ($a_0,a^*$), previous weights $W$, $k_w$-bit, $k_a$-bit, shift threshold parameter $\xi$ ($\epsilon_w$) and learning rate $\eta$.\\
\specialrule{0em}{1pt}{1pt}
\textbf{Ensure} Updated weights $W$.\\
\specialrule{0em}{1pt}{1pt}
\specialrule{0em}{1pt}{1pt}
\ \ {\textbf{1. Computing the parameters gradients}:}\\
\ \ {\textbf{1.1. Forward propagation}:}\\
\ \ \ \ \textbf{for} $i$=1 to $L$ \textbf{do},\\
\ \ \ \ \ \ $T_i \gets Q_{k_w}(W_i,\epsilon_w)$ with \eqref{eq:quantization_weight}\\
\ \ \ \ \ \ Compute $\alpha$ with \eqref{eq:linear_regression_problem}\\
\ \ \ \ \ \ $s_i \gets A^q_{i-1}\alpha T_i$\\
\ \ \ \ \ \ Apply max-pooling\\
\ \ \ \ \ \ $a_i \gets$ BatchNorm($s_i$)\\
\ \ \ \ \ \ \textbf{if} $i<L$ \textbf{then}\\
\ \ \ \ \ \ \ \ $A^q_i \gets Q_{k_a}(a_i,\epsilon_a)$ with \eqref{eq:quantizaiton_activation}\\
\specialrule{0em}{2pt}{2pt}
\ \ {\textbf{1.2. Backward propagation}:}\\
\ \ \{note that the gradients are full-precision.\}\\
\ \ Compute $g_{a_L}=\frac{\partial \mathcal{L}}{\partial a_L}$\ knowing $a_L$ and $a^*$\\
\ \ \textbf{for} $i=L$ to 1 \textbf{do}\\
\ \ \ \ \textbf{if} $i<L$ \textbf{then}\\
\ \ \ \ \ \ $g_{a_i} \gets g_{A^q_i} \otimes 1_{|a_i|\leqslant0.5}$ by Algorithm 2\\
\ \ \ \ \textbf{end if}\\
\ \ \ \ $g_{s_i} \gets$ BackBatchNorm($g_{a_i},s_i$)\\
\ \ \ \ $g_{A^q_{i-1}} \gets g_{s_i}T_i$\\
\ \ \ \ $g_{T_i} \gets g_{s_i}^\top A^q_{i-1}$\\
\ \ \textbf{end for}\\
\specialrule{0em}{2pt}{2pt}
\ \ {\textbf{2. Accumulating the parameters gradients}:}\\
\ \ \textbf{for} $i=1$ to $L$ \textbf{do}\\
\ \ \ \ With $g_{T_i}$ known, compute $g_{W_i}$ by Algorithm 1\\
\ \ \ \ $W_i \gets$ Update($W_i,\eta,g_{W_i}$)\\
\ \ \ \ $\eta \gets \lambda\eta$\\
\ \ \textbf{end for}\\
\specialrule{0em}{2pt}{2pt}
\hline
\end{tabular}
\label{Algorithm 3}
\end{table}

\section{Result and Discussion}\label{sec:Result_and_Discussion}
In this section, the proposed quantization method and fusion strategies are evaluated on three famous benchmark datasets. The following are considered: 1) the proposed dynamic fusion models are compared with other baseline models, 2) the effect of weight shift threshold parameter is evaluated, 3) the trade-off between quantization and model accuracy. For the first method (we call it Baseline or (TerNet) method), the required sensor signal sources are stacked together. In the second method (referred to as FTerNet), different sensor signal sources are processed through their own sub-networks and fused together with the learned representations before the dense layer, i.e., each element of fusion weights is equal to 1. The model proposed in this paper (DFTerNet), differs from the second method discussed in the way it handles the fusion part. In the DFTerNet, each element of fusion weights is sampled from a Bernoulli distribution given by the scale parameter of the quantization method that we proposed.

\subsection{Multi-sensor Signals Fusion}\label{subsec:MSF}
In order to evaluate the different fusion strategies which were described in Section \ref{sec:fusion_stragegy}, an ablation study was performed on the OPPORTUNITY and PAMAP2 datasets. The first set of experiments consisted of comparing the three fusion strategies on the each dataset. As shown by the \textbf{bold scores} in Table \ref{tab:overall_performance_comparison:_OPPORTUNITY_and_UniMiB-SHAR}, the order of fusion performance is: \textbf{matched dynamic fusion} in first place, followed by \textbf{late fusion} and finally \textbf{early fusion}. The reason for this is that it is better for each sensor signal source to have its network but improper to apply a single network to unify all signal sources. Meanwhile, there is a correlation between different signal sources and activity types and therefore the result of the recognition should be more reliable when the signal sources are highly correlated with the activity type. According to the two points above, the recognition result should be weighted by the learned representations of multiple signal sources, and the weight of learned representations of each signal source should reflect the similarity between the signal source and the activity type.

\begin{table}[tp]
\centering
\begin{threeparttable}
  \centering
  \caption{Comparison of $\xi$'s value for the activity recognition performances (Weighted $F1$-score) on OPPORTUNITY dataset.}
  \label{tab:xisvalue}
    \begin{tabular}{cccccc}
    \toprule
    $\xi$ &=2.6&2.7&\textbf{2.8}&\textbf{2.9}&3.0\cr
    \midrule
    \specialrule{0em}{2pt}{2pt}
    DFTerNet ($\boldsymbol{g}_p$)&0.884&0.897&\textbf{0.910}&\textbf{0.909}&0.893\cr
    \specialrule{0em}{2pt}{2pt}
    DFTerNet ($\boldsymbol{g}_s$)&0.879&0.894&\textbf{0.905}&\textbf{0.905}&0.891\cr
    \bottomrule
    \end{tabular}
\end{threeparttable}
\end{table}

\subsection{Analysis of Weight Shift Threshold Parameter}
In our DFTerNet, the result of the weight shift parameter $\xi$ of the $Q_w(\cdot)$ operation will directly affect the following fusion weights $\varphi$. Therefore, the second set of experiments consider the effect of $\xi$'s value. As mentioned in the previous section, the value of $\epsilon_w$ is related to the value of $\xi$ and the fusion weights $\varphi$ are sampled from $Bernoulli(Q_{k_w}(W,\epsilon_w),m)$. We use \textbf{matched dynamic fusion}-$\boldsymbol{g}_p$ and \textbf{matched dynamic fusion}-$\boldsymbol{g}_s$ on the OPPORTUNITY dataset as a test case to compare the performance of $\xi$'s value. In this experiment, the parameter settings are the same as described in Section \ref{sec:Experimental_Setup} and Section \ref{sec:Implementation_Details}. Table \ref{tab:xisvalue} summarizes the results of $\xi$'s value on \textbf{matched dynamic fusion}. It can be seen that the quantization method we proposed achieves its best performance when using $\xi=2.8$ or $\xi=2.9$. Similar phenomenon can also be found in~\cite{b13}.

\begin{table*}
\caption{(a) Weighted $F_1\text{-score}$ performances of different fusion strategies, (F)BNNs (1-bit) and our proposed models (2-bit) for activity recognition on the \textbf{\underline{O}}PPORTUNITY, \textbf{\underline{P}}AMAP2 and \textbf{\underline{U}}niMiB-SHAR datasets. (b) $\mathbf{\epsilon}$-$\mathbf{Q}$ quantization method generates 2-bit weight and activation Convolutional networks for activity recognition with the ability to make faithful inference and roughly $11\times$ fewer parameters than its counterpart.}
\label{tab:double_sub_figure}
\centering
\subtable[]{
    \begin{threeparttable}
    \label{tab:overall_performance_comparison:_OPPORTUNITY_and_UniMiB-SHAR}
    \begin{tabular}{ccccc}
    \toprule
    \specialrule{0em}{2pt}{2pt}
    \multirow{2}{*}{\small{Method}}&
    \multicolumn{2}{c}{O}&\multicolumn{1}{c}{P}&\multicolumn{1}{c}{U}\cr
    \cmidrule(lr){2-3}\cmidrule(lr){4-4}\cmidrule(lr){5-5}
    &locomotion$\dagger$&gestures$\ddagger$&Activities$\dagger$&ADLs and falls\cr
    \specialrule{0em}{3pt}{3pt}
    \midrule
    \specialrule{0em}{3pt}{3pt}
    Early fusion ~\cite{b44}&0.876 $\pm$0.09&0.881 $\pm$0.11&0.867 $\pm$0.09&0.7981 $\pm$0.12\cr
    \specialrule{0em}{1pt}{1pt}
    BNN~\cite{b16}&0.752 $\pm$0.17&0.751 $\pm$0.20&0.733 $\pm$0.14&0.6481 $\pm$0.21\cr
    \specialrule{0em}{1pt}{1pt}
    TerNet (Early fusion)&0.865 $\pm$0.14&0.876 $\pm$0.19&0.850 $\pm$0.11&0.7727 $\pm$0.20\cr
    \specialrule{0em}{3pt}{3pt}
    \hline
    \specialrule{0em}{3pt}{3pt}
    Late fusion ~\cite{b45}&0.897 $\pm$0.06&0.917 $\pm$0.05&0.908 $\pm$0.05&\textbf{---}\cr
    \specialrule{0em}{1pt}{1pt}
    FBNN~\cite{b16}&0.765 $\pm$0.16&0.773 $\pm$0.19&0.764 $\pm$0.15&\textbf{---}\cr
    \specialrule{0em}{1pt}{1pt}
    FTerNet (Late fusion)&0.883 $\pm$0.10&0.908$\pm$0.14&0.893 $\pm$0.13&\textbf{---}\cr
    \specialrule{0em}{3pt}{3pt}
    \hline
    \specialrule{0em}{3pt}{3pt}
    Dynamic fusion-$\boldsymbol{g}_p$$\dagger$&\textbf{0.915} $\pm$0.04&\textbf{---}&\textbf{0.914} $\pm$0.06&\textbf{---}\cr
    \specialrule{0em}{1pt}{1pt}
    DFTerNet ($\boldsymbol{g}_p$)$\dagger$&\textbf{\emph{0.909}} $\pm$0.06&\textbf{---}&\textbf{\emph{0.901}} $\pm$0.11&\textbf{---}\cr
    \specialrule{0em}{3pt}{3pt}
    \hline
    \specialrule{0em}{3pt}{3pt}
    Dynamic fusion-$\boldsymbol{g}_s$$\ddagger$&\textbf{---}&\textbf{0.920} $\pm$0.07&\textbf{---}&\textbf{---}\cr
    \specialrule{0em}{1pt}{1pt}
    DFTerNet ($\boldsymbol{g}_s$)$\ddagger$&\textbf{---}&\textbf{\emph{0.910}} $\pm$0.10&\textbf{---}&\textbf{---}\cr
    \specialrule{0em}{3pt}{3pt}
    \bottomrule
    \end{tabular}
        \begin{tablenotes}
        \item[$\dagger$] Dynamic fusion-$\boldsymbol{g}_p$ \textbf{match} the locomotion category of OPPORTUNITY and PAMAP2.
        \item[$\ddagger$] Dynamic fusion-$\boldsymbol{g}_s$ and the gestures category are \textbf{matched}.
        \end{tablenotes}
    \end{threeparttable}
}
\qquad
\subtable[]{
  \begin{threeparttable}
  \label{tab:overall_performance_comparison:_Memory}
  \centering
    \begin{tabular}{cc|c}
    \toprule
    & & Memory\cr
    \midrule
    \specialrule{0em}{1.5pt}{1.5pt}
    \multirow{3}*{(F)BNN}&O&$\sim$20k\\
                         &P&$\sim$10k\\
                         &U&$\sim$0.9k\\
    \specialrule{0em}{1.8pt}{1.8pt}
    \hline
    \specialrule{0em}{1.8pt}{1.8pt}
    \multirow{3}*{TerNet}&O&$\sim$39k\\
                         &P&$\sim$20k\\
                         &U&$\sim$1.8k\\
    \specialrule{0em}{1.8pt}{1.8pt}
    \hline
    \specialrule{0em}{1.8pt}{1.8pt}
    \multirow{3}*{FTerNet}&O&$\sim$40k\\
                         &P&$\sim$24k\\
                         &U&$\sim$1.9k\\
    \specialrule{0em}{1.8pt}{1.8pt}
    \hline
    \specialrule{0em}{1.8pt}{1.8pt}
    \multirow{3}*{DFTerNet}&O&$\sim$\textbf{34k}\\
                         &P&$\sim$\textbf{17k}\\
                         &U&$\sim$\textbf{1.8k}\\
    \specialrule{0em}{1.8pt}{1.8pt}
    \hline
    \specialrule{0em}{1.8pt}{1.8pt}
    \multirow{3}*{FP}&O&$\sim$0.38M\\
                     &P&$\sim$0.22M\\
                     &U&$\sim$17k\\
    \specialrule{0em}{1.5pt}{1.5pt}
    \bottomrule
    \end{tabular}
    \end{threeparttable}
}
\end{table*}

\subsection{Visualization of The Quantization Weights}
\begin{figure}
  \centering
  \includegraphics[width=8cm]{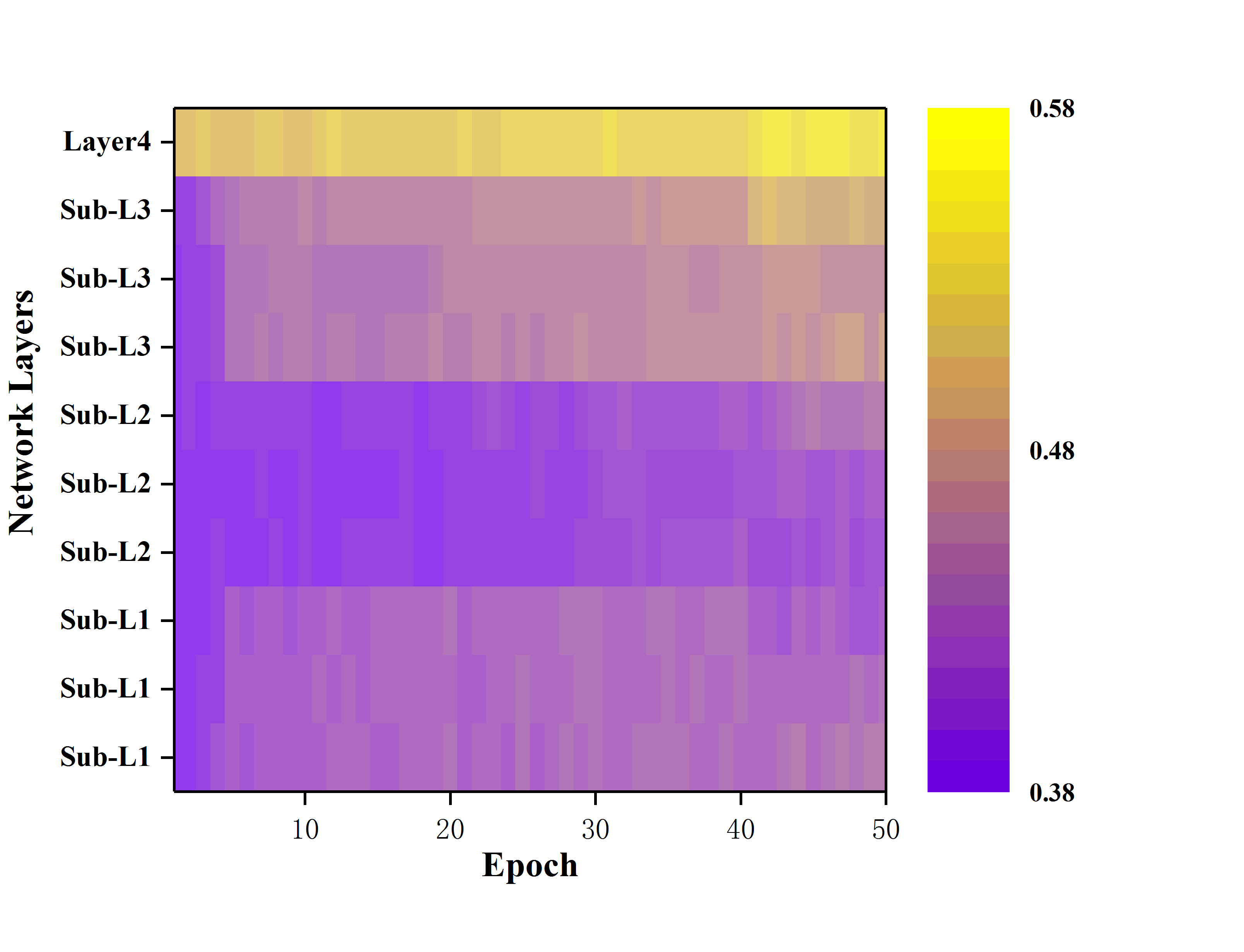}\\
  \caption{Visualization of fraction of \textbf{zero value} at each epoch in DFTerNet ($\boldsymbol{g}_p$) on the locomotion category of OPPORTUNITY dataset.}\label{fig:F4}
\end{figure}

In addition to analyzing the quantized weights, we further looked inside the learned layers and checked the values.
We plot the heatmap of the fraction of zero value by DFTerNet-($\boldsymbol{g}_p$) on the locomotion category of the OPPORTUNITY dataset across epochs. As shown in Figure \ref{fig:F4}, we can see that the fraction of zero values increase in later epochs, similar phenomena also appear in DFTerNet-($\boldsymbol{g}_p$) on PAMAP2 dataset and DFTerNet-($\boldsymbol{g}_s$) on the gestures category of the OPPORTUNITY dataset. Section \ref{sec:proof_section} proves the reconstruction error $e$ boundary, the model can achieve a very small $e$ value. Table \ref{tab:computational} shows that the layers contain most of the free parameters with increased sparsity at the end of training, this indicates that our proposed quantization method can avoid overfitting and sparsity acts as a regularizer.

\begin{figure*}
  \centering
  \includegraphics[width=17.5cm]{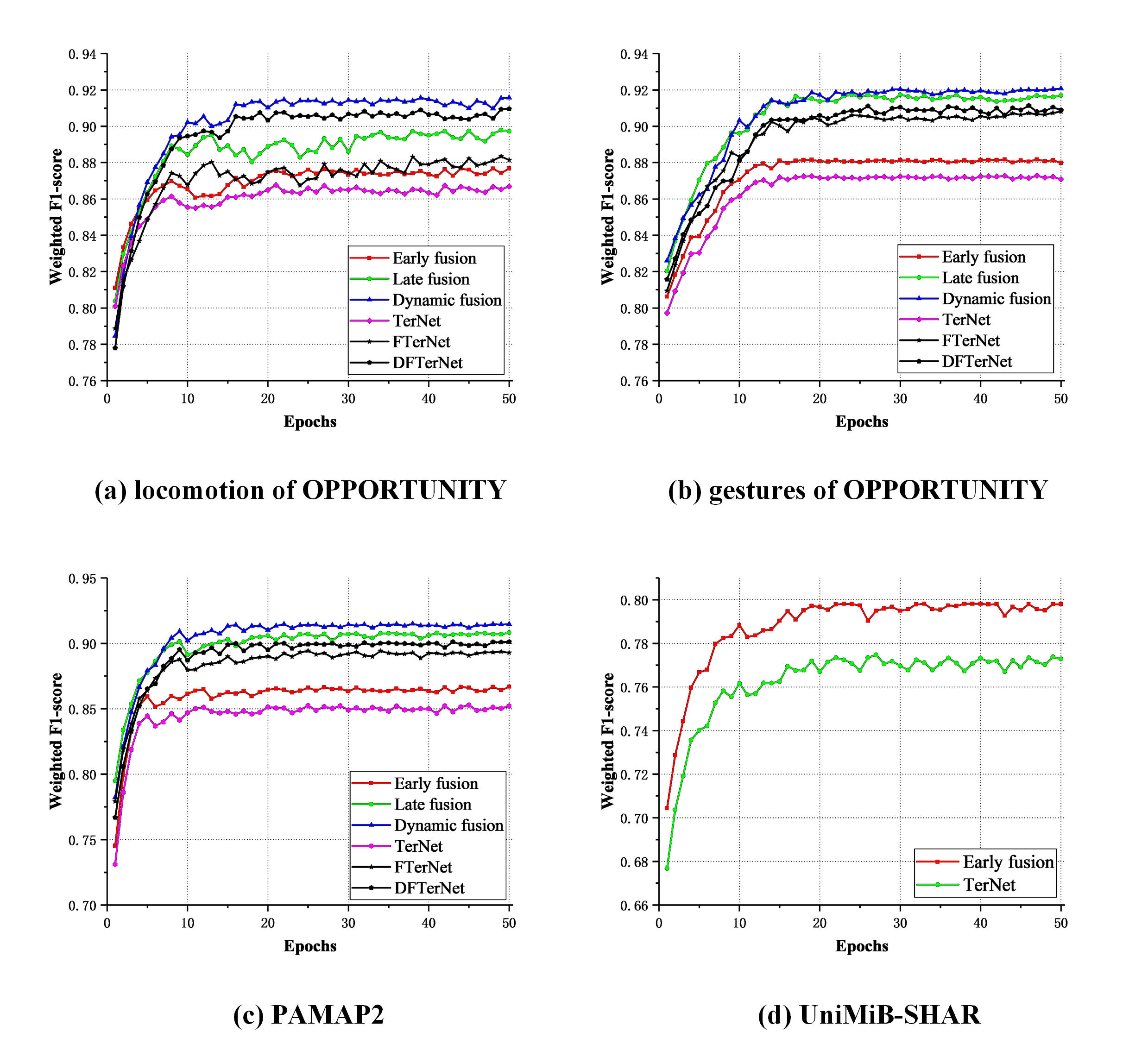}\\
  \caption{Validation Weighted $F_1$-score Curves on These Datasets.}\label{fig:F5}
\end{figure*}

\subsection{The Trade-off Between Quantization and Model Accuracy}
The third set of experiments are performed to explore the model accuracy of the quantization method. Just like in the first and second sets of experiments, a four-layer convolutional network is used and the parameter settings for the sliding window as well as batch size are kept completely the same. The weight shift threshold parameter $\xi$ is set to $\xi=2.8$. Finally, TerNet, FTerNet and DFTerNet with their own full-precision counterparts are generated for comparison. Table \ref{tab:double_sub_figure} shows the weighted $F_1$-score performance of different full-precision models and their counterparts, which are described in Figure \ref{fig:F2}. It also depicts the memory usage of all models mentioned above. Table \ref{tab:overall_performance_comparison:_OPPORTUNITY_and_UniMiB-SHAR} shows that using the proposed quantization method, results in a very small difference in performance between TerNet (2-bit) and FTerNet (2-bit) network and its full-precision counterpart, while beating BNN (1-bit) and FBNN (1-bit) by a large margin. On the fewer channels (i.e., 3-channels) UniMiB-SHAR dataset, BNN and TerNet both get poorer performance than the full-precision counterpart. However, the accuracy gap between TerNet and the full-precision counterpart is smaller than the gap between BNN and the full-precision counterpart. Thus, TerNet beat BNN again.

Figure \ref{fig:F5} shows the validation weighted $F_1$-score curves on these datasets. As shown in the Figure \ref{fig:F5}, our quantized models (TerNet, FTerNet and DFTerNet) converge almost as fast and stably as their counterparts. This demonstrates the robustness of the quantization technique we proposed.

The efficiency of using Hamming distance calculation by \eqref{eq:our_Hamming_space} is another indicator of this work, compared with (16/32)-bit floating point, 2-bit (even 8-bit) operations will not only reduce the energy costs for Hardware/IC Design (see Figure~\ref{fig:F1}), but also halves the memory access costs and memory size requirements during \textbf{run-time}, which will greatly facilitate the deployment of binary/ternary convolutional neural networks on portable devices~\cite{b55}. For example, training a dynamic fusion model using the OPPORTUNITY dataset took about $\sim$12 minutes on an NVIDIA TITAN Xp 12G GPU test platform. For inference the full-precision network on a CPU takes about 15 seconds. We estimate that the DFTerNet inference time to be $\sim$1.8 seconds on a mobile CPU. This shows that the quantization technique we proposed can achieve a $\sim9\times$ speedup.

\section{Conclusion and Future Work}\label{sec:Conclusion_and_Future_Work}
In this paper, we present DFTerNet, a new network quantization method and a novel dynamic fusion strategy, to address the problem of how to better recognize activities from multi-sensor signal sources and deploy them on low-computation capable portable devices. Firstly, the proposed quantization method $\boldsymbol{\epsilon }$-$\boldsymbol{Q}$ is called by two operations through adjusting the scale parameter $\epsilon$, weight quantization $Q_{k_w}(W,\epsilon_w)$ and activation quantization $Q_{k_a}(A,\epsilon_a)$. Secondly, the bit-counts scheme that replaces matrix multiplication proposed in this work is hardware friendly and realizes a $\sim$9$\times$ speedup as well as requiring $\sim$11$\times$ less memory. Thirdly, a novel dynamic fusion strategy was proposed. Unlike existing methods which treat the representations from different sensor signal sources equally, it considers the fact that different sensor signal sources need to be learned separately and less ``contribution'' signal sources reduce its representations by fusion weights which are sampled from a Bernoulli distribution given by the $Q_{k_w}(W,\epsilon_w)$. Experiments that were performed demonstrated the effectiveness of the proposed quantization method and dynamic fusion strategy. As for future works, we plan to extend the quantization method to quantization gradients and errors so that it can be deployed directly on portable devices for training and inference. Because improvement of model performance requires continuous online learning, separation of training and inference will limit that.

\section*{Acknowledgment}

The authors would also like to thank the associate editor and anonymous reviewers for their comments to improve the paper.


\end{document}